# SDTN and TRN: Adaptive Spectral-Spatial Feature Extraction for Hyperspectral Image Classification

**Fuyin Ye**
*Wenzhou Agricultural Science Research Institute*
Wenzhou, China
yefuyin@wzvcst.edu.cn

*Erwen Yao*
*Wenzhou Agricultural Science Research Institute*
Wenzhou, China
yaoerwen@wzvcst.edu.cn

**Jianyong Chen**
*Wenzhou Agricultural Science Research Institute*
Wenzhou, China
chenjianyong@wzvcst.edu.cn

*Fengmei He*
*Wenzhou Agricultural Science Research Institute*
Wenzhou, China
hefengmei@wzvcst.edu.cn

**Junxiang Zhang**
*Wenzhou Agricultural Science Research Institute*
Wenzhou, China
zhangjunxiang@wzvcst.edu.cn

Lihao Ni*
*Wenzhou Agricultural Science Research Institute*
Wenzhou, China
nilihao@wzvcst.edu.cn

*Abstract*—Hyperspectral images (HSIs) classification has become a foundational tool in the field of precision agriculture, enabling the extraction of detailed information critical for applications such as monitoring crop conditions, identifying plant diseases, and assessing soil characteristics. Nevertheless, conventional approaches are often constrained by the inherently high dimensionality of hyperspectral data, redundancy across spectral-spatial domains, and the limited availability of annotated training samples, resulting in less-than-optimal performance. In response to these limitations, we propose a novel Self-Adaptive Tensor-Regularized Network (SDTN), which integrates tensor decomposition with regularization strategies, allowing for dynamic adjustment of tensor rank in accordance with data complexity. This adaptive mechanism facilitates more expressive and compact feature representations. Expanding upon SDTN, we further introduce the Tensor-Regularized Network (TRN), a lightweight structure that incorporates the spectral-spatial features derived from SDTN and enhances them through multi-scale processing. TRN not only achieves high classification accuracy but also reduces computational complexity, making it highly suitable for real-time deployment in resource-constrained conditions. Experiments on the PaviaU benchmark datasets demonstrate improvements in accuracy and reduced model parameters compared to state-of-the-art methods.

*Keywords—Tensor decomposition, Hyperspectral images(HSIs) classification, Spectral-Spatial features*

## I. INTRODUCTION

Hyperspectral images(HSIs) classification is important in precision agriculture, giving critical support for crop health monitoring, early disease detection, soil analysis and so on. Compared with traditional RGB images, HSIs contain hundreds of continuous spectral bands, offering richer information that enables subtle distinctions between crop features and soil properties.

traditional hyperspectral classification methods, such as Support Vector Machines (SVM) [1] and K-Nearest Neighbors (KNN) [2], primarily relied on manually designed features, limiting capability to capture intricate spectral-spatial relationships [3]. Recently, deep learning frameworks, particularly Convolutional Neural Networks (CNNs), have significantly enhanced classification performance by automatically learning relative spectral-spatial features. For example, Sun et al. [3] proposed a 3D-2D mixed CNN combined with a attention-free transformer, achieving promising results for hyperspectral classification tasks. Similarly, Ye et al. [4] introduced a Hybrid Spatial-Spectral Tensor Network(HSSTN) that effectively extracts low-rank features. Nevertheless, CNN-based methods still face difficulties such as data redundancy due to high dimensionality, high computational complexity, and susceptibility to overfitting, especially with limited labeled samples.To address these issues, tensor decomposition techniques such as Tucker and CP decomposition have been introduced to compactly represent hyperspectral data [5]. However, these methods typically use fixed tensor ranks, lacking adaptability, and thus often lead to either overfitting or omission of crucial spectral-spatial information [6].

Consequently, it remains challenging to simultaneously tackle high-dimensional data redundancy, limited labeled samples, and computational efficiency, particularly for real-time deployment on resource-constrained edge devices. To address these challenges, we first propose the Self-Adaptive Tensor-Regularized Network (SDTN), which adaptively captures robust spectral-spatial features by dynamically adjusting tensor ranks according to data complexity. Building upon SDTN, we further propose the Tensor-Regularized Network (TRN), a lightweight framework incorporating tensor regularization to refine extracted features, effectively balancing classification accuracy with computational efficiency, and thus enabling practical real-time deployment.

The main contributions of this study include:

**(1) A novel SDTN framework:** Adaptively extracts robust spectral-spatial features via dynamic tensor decomposition, effectively addressing redundancy and overfitting.

**(2) A lightweight TRN framework:** Integrates SDTN-extracted features with tensor-regularized modules, achieving an optimal balance between classification accuracy and computational efficiency.

**(3) Comprehensive evaluation on the PaviaU dataset:** Extensive experiments demonstrate that the proposed SDTN

*Corresponding Author: Lihao Ni, Ph.D. Candidate, specializing in Artificial Intelligence and Image Recognition.
This work is supported by Zhejiang Province 14th Five-Year Plan Teaching Reform Project (Grant No.jg20230464) and Scientific Research Fund of Zhejiang Provincial Education Department (Grant No.Y202456297) and TRN frameworks outperform existing methods, validating their effectiveness in hyperspectral image classification tasks.

## II. PROPOSED METHOD

In this section, as illustrated in Fig 1, we propose the Self-Adaptive Tensor-Regularized Network (SDTN) and the Tensor-Regularized Network (TRN).

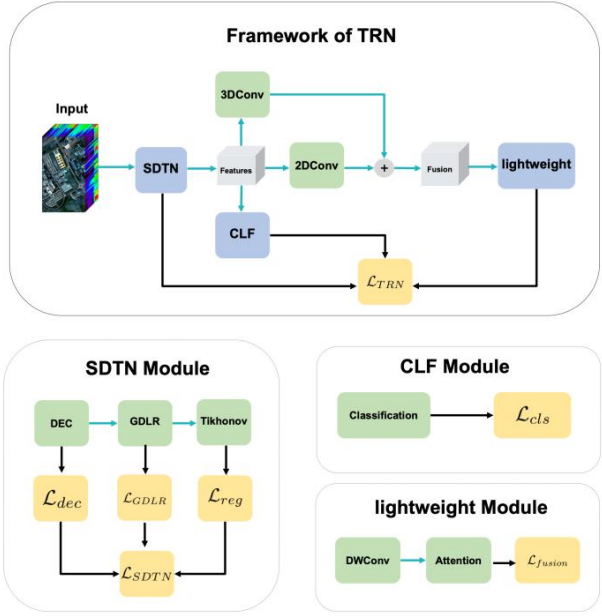

Fig. 1. The framework of TRN

### A. Self-Adaptive Tensor-Regularized Network (SDTN)

Hyperspectral imagery inherently exhibits high dimensionality and complex spectral-spatial correlations, posing significant challenges to conventional tensor modeling approaches. To effectively model these complex intrinsic relationships and overcome the limitations of traditional tensor decomposition methods, we propose a novel Self-Adaptive Tensor-Regularized Network (SDTN). This framework adaptively captures intrinsic spectral-spatial correlations through a fully connected tensor-based decomposition with gradient-domain low-rank matrix constraints and Tikhonov regularization, significantly enhancing the robustness and discriminative power of the spectral-spatial representations.

Given a hyperspectral image patch represented by a high-order tensor $\mathcal{X} \in \mathbb{R}^{I_1 \times I_2 \times \cdots \times I_N}$, SDTN adaptively decomposes it into a series of interconnected tensor factors. Unlike conventional decomposition models, SDTN explicitly models complex inter-modal correlations through an adaptive fully-connected tensor structure:

$$\mathcal{X}(i_1, i_2, \ldots, i_N) = \sum_{r_{1,2}=1}^{R_{1,2}} \cdots \sum_{r_{N-1,N}=1}^{R_{N-1,N}} \prod_{k=1}^{N} \mathcal{G}_k(\mathbf{r}_k, i_k) \quad (1)$$

Where $\mathcal{G}_k$ denotes tensor factors associated with each mode, designed to adaptively capture intrinsic structural relationships without explicitly referencing conventional fully-connected tensor decomposition approaches. Such adaptive modeling significantly enhances feature representation capabilities by capturing more comprehensive inter-modal correlations.

To better model local spatial continuity and spectral smoothness intrinsic to hyperspectral imagery, SDTN introduces an innovative gradient-domain low-rank constraint to the mode-unfolded tensor factors. Specifically, for $\mathcal{G}_k^{(k)}$ (the unfolded matrix), the constraint is expressed as:

$$D_k G_k^{(k)} = U_k V_k \quad (2)$$

where $D_k$ is a gradient operator along the k-th mode, capturing local continuity and spectral correlations. The low-rank matrices $U_k \in \mathbb{R}^{I_k \times r_k}$ and $V_k \in \mathbb{R}^{r_k \times \Pi_{j \neq k} R_{k,j}}$ enforce concise representations, reducing redundancy, and enhancing the robustness of the extracted features.

To further stabilize the adaptive tensor decomposition and prevent overfitting, particularly under conditions with limited labeled data, we incorporate Tikhonov regularization as follows:

$$\mathcal{L}_{reg} = \sum_{k=1}^{N} \left( \lambda_1 \|U_k\|_F^2 + \lambda_2 \|V_k\|_F^2 + \lambda_3 \|\mathcal{G}_k\|_F^2 \right) \quad (3)$$

This regularization effectively controls the complexity of the model, improves numerical stability, and significantly enhances generalization ability on unseen data.

To explicitly guide feature extraction toward classification, we incorporate supervised classification loss based on labeled samples. Let $f(\mathcal{G}_k)$ be the predicted classification probabilities obtained through a classifier layer, and $y$ the true labels. The classification loss is defined as following:

$$\mathcal{L}_{cls}(f(\mathcal{G}_k), y) = -\sum_{i=1}^{M} y_i \log f(\mathcal{G}_k)_i \quad (4)$$

Here, $M$ denotes the number of classes, ensuring the learned tensor features directly contribute to discriminative classification.

The complete optimization objective integrating all components—reconstruction, gradient-domain low-rank constraint, Tikhonov regularization, and supervised classification—is defined as follows:

$$\mathcal{L}_{SDTN} = \frac{1}{2} \|\mathcal{X} - \mathcal{F}(\{\mathcal{G}_k\})\|_F^2 + \frac{\alpha}{2} \sum_{k=1}^{N} \|D_k \mathcal{G}_k^{(k)} - U_k V_k\|_F^2 + \mathcal{L}_{reg} + \beta \mathcal{L}_{cls} \quad (5)$$

These innovative components collectively establish SDTN as a powerful and novel framework: adaptive fully-connected tensor modeling captures comprehensive inter-modal correlations beyond traditional adjacent mode limitations, significantly enhancing representation capability; the novel gradient-domain low-rank constraint effectively models spatial continuity and spectral smoothness, substantially reducing redundancy and increasing robustness to spectral noise; and the incorporation of Tikhonov regularization significantly improves numerical stability and generalization performance, particularly beneficial under limited labeled data conditions. Consequently, SDTN achieves superior spectral-spatial feature extraction and classification accuracy, making it highly suitable for hyperspectral image classification tasks in precision agriculture and related applications.

*B. Tensor-Regularized Network (TRN)*

While the Self-Adaptive Tensor-Regularized Network (SDTN) excels in robustly extracting spectral-spatial features through adaptive tensor decomposition and regularization strategies, practical agricultural scenarios demand models optimized specifically for lightweight edge-device deployment. To address this need without compromising the strong feature representation capabilities of SDTN, we propose a complementary model, the Tensor-Regularized Network (TRN). As illustrated in Figure 1, TRN effectively integrates the discriminative tensor-based features derived from SDTN into a computationally efficient network architecture, ensuring an optimal balance between classification performance and computational efficiency.

In TRN, the spectral-spatial feature representation provided by SDTN is explicitly obtained through tensor reconstruction from optimized tensor factors:

$$\mathcal{H}(i_1, i_2, \ldots, i_N) = \sum_{r_{1,2}=1}^{R_{1,2}} \cdots \sum_{r_{N-1,N}=1}^{R_{N-1,N}} \prod_{k=1}^{N} \mathcal{G}_k(\mathbf{r}_k, i_k) \quad (6)$$

where $\mathcal{H}(i_1, i_2, \ldots, i_N)$ denotes the reconstructed tensor feature, encapsulating essential spectral-spatial information.

Furthermore, we design TRN as an end-to-end simultaneously optimized architecture, allowing the SDTN tensor decomposition and CNN-based convolutional feature extraction to be jointly trained. Specifically, TRN employs dual convolutional pathways directly utilizing the tensor features from SDTN:

**3D Convolutional Pathway:**

$$F_{3D} = Conv3D(\mathcal{H}, W_{3D}) + b_{3D} \quad (7)$$

**2D Convolutional Pathway:**

$$F_{2D} = Conv2D(\text{reshape}(\mathcal{H}), W_{2D}) + b_{2D} \quad (8)$$

The outputs from the 3D and 2D convolutional pathways are fused together:

$$F_{fusion} = Concat(F_{3D}, F_{2D}) \quad (9)$$

Considering the strict resource constraints of practical agricultural applications (e.g., drones or edge computing terminals), we further optimize model's structure through lightweight designs, explicitly employing depthwise separable convolutions and channel-wise attention mechanisms.

**Depthwise Separable Convolutions:**

$$Conv_{dw}(F_{fusion}) = Conv_{pointwise}(Conv_{depthwise}(F_{fusion})) \quad (10)$$

**Channel-wise Attention Mechanism:**

$$F_{attention} = Sigmoid(FC(ReLU(FC(Conv_{dw}(F_{fusion}))))) \odot Conv_{dw}(F_{fusion}) \quad (11)$$

These optimizations significantly reduce computational complexity and memory usage, making TRN highly suitable for real-time inference on edge devices.

To ensure comprehensive collaboration between SDTN's tensor representation and TRN's classification performance, we define a unified optimization objective integrating reconstruction accuracy, gradient-domain low-rank constraints, Tikhonov regularization, supervised classification loss, and tensor consistency constraint:

$$\mathcal{L}_{TRN} = \frac{1}{2}\|\mathcal{X} - \mathcal{F}(\{\mathcal{G}_k\})\|_F^2 + \frac{\alpha}{2}\sum_{k=1}^{N}\|D_k\mathcal{G}_k^{(k)} - U_kV_k\|_F^2 + \mathcal{L}_{reg} + \beta\mathcal{L}_{cls}(y, \hat{y}) + \gamma\|\mathcal{H} - F_{fusion}\|_F^2 \quad (12)$$

This explicit end-to-end collaborative training ensures TRN attains robust, efficient, and discriminative tensor features ideal for precision agriculture.

## III. EXPERIMENT

*A. SettingDataset*

We evaluate our method on the Pavia University (PaviaU) hyperspectral dataset. The image, captured by the ROSIS sensor, consists of 610 × 340 pixels and 103 spectral channels (after noise removal), with a spatial resolution of 1.3 m. The dataset contains 9 land-cover classes and 42,776 labeled samples.

*B. Setting*

To assess performance, we randomly selected 10 labeled samples per class for training. The proposed SDTN and TRN were compared with two state-of-the-art models (SSATN [7], SRSN [8]), an ablation variant (CNN), and classical baselines (SVM [1], KNN [2]). Evaluation metrics include overall accuracy (OA), average accuracy (AA), and the Kappa coefficient [9]. All models were implemented in TensorFlow 1.10.0 and trained on a workstation with an Intel Xeon E5-2650 v4 CPU, 64 GB RAM, and an NVIDIA GTX 1080Ti GPU. The learning rate started at 0.001 and decayed by 0.9 every 10,000 iterations.

*C. Performance Evaluation*

As shown in Table I and Fig. 2, the proposed SDTN and TRN frameworks consistently outperform both classical baselines and state-of-the-art methods on the PaviaU dataset.

SDTN achieves a 7.08% improvement in overall accuracy over the ablation baseline CNN, demonstrating the effectiveness of its adaptive tensor decomposition in capturing global spectral-spatial correlations under limited supervision. It also shows strong generalization in spectrally similar regions.

Built upon SDTN, TRN further improves all metrics, achieving the highest OA (99.72%), AA (99.81%), and Kappa (99.35%) among all compared methods. Its lightweight tensor-regularized architecture enhances feature representation while maintaining computational efficiency.

The classification maps in Fig. 2 show that both models produce cleaner and more coherent predictions. SDTN effectively reduces scattered errors, while TRN further improves boundary sharpness and spatial consistency, highlighting their complementary strengths in hyperspectral image classification.

TABLE I.     CLASSIFICATION RESULTS OF DIFFERENT METHODS ON PAVIAU (THE BEST RESULTS ARE IN BOLD)

| Class | SVM | KNN | CNN | SRSN | SSATN | SDTN | TRN |
|---|---|---|---|---|---|---|---|
| 1 | 60.19 | 66.61 | 94.53 | 99.44 | 99.73 | 99.74 | **99.97** |
| 2 | 55.01 | 62.66 | 93.55 | 99.20 | **99.99** | 99.80 | **99.99** |
| 3 | 67.4 | 44.57 | 97.08 | 97.03 | 91.91 | 98.09 | **100** |
| 4 | 85.56 | 89.03 | 88.80 | **100** | 99.34 | **100** | **100** |
| 5 | 77.75 | 99.33 | 100 | **100** | 99.93 | **100** | **100** |
| 6 | 45.57 | 53.56 | 99.54 | 88.2 | 94.20 | 96.81 | **100** |
| 7 | 92.95 | 91.67 | 96.21 | **100** | **100** | **100** | **100** |
| 8 | 46.60 | 66.97 | 64.98 | 96.19 | 97.82 | 98.53 | **99.70** |
| 9 | 99.89 | 99.89 | 99.89 | **100** | **100** | **100** | **100** |
| OA | 59.64 | 66.44 | 92.20 | 97.71 | 98.7 | 99.28 | **99.72** |
| AA | 70.10 | 74.92 | 92.73 | 97.79 | 98.18 | 99.22 | **99.81** |
| Kappa | 50.13 | 58.05 | 89.77 | 96.95 | 98.27 | 99.04 | **99.35** |

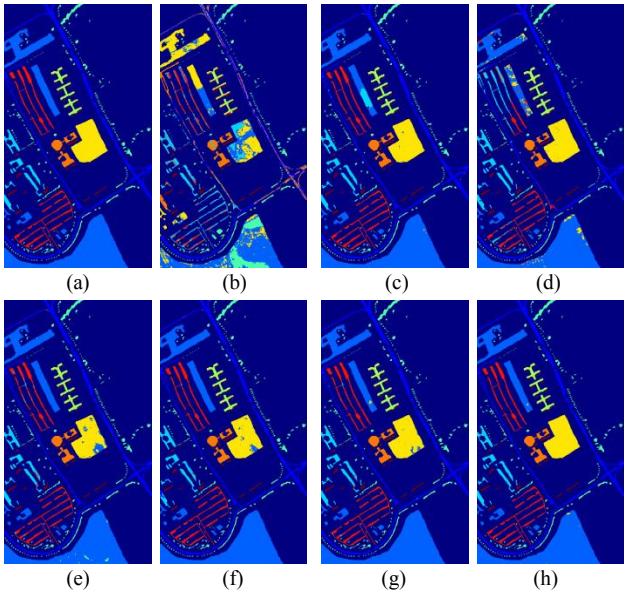

(a)    (b)    (c)    (d)
(e)    (f)    (g)    (h)

Fig. 2. Classification maps of different methods on the PaviaU dataset: (a) Ground truth. (b) SVM. (c) KNN. (d) CNN. (e) SRSN. (f) SSATN. (g) SDTN. (h) TRN.

## IV. CONCLUSION

In this paper, we first propose the Self-Adaptive Tensor-Regularized Network (SDTN), which utilizes adaptive tensor decomposition to robustly extract spectral-spatial features, effectively addressing high-dimensional data redundancy. Building upon SDTN, we further propose the Tensor-Regularized Network (TRN), a lightweight framework that significantly improves classification accuracy and computational efficiency by incorporating tensor-regularized structures. Experimental results demonstrate superior performance of our methods over state-of-the-art approaches, highlighting their practicality for real-time hyperspectral image classification tasks in resource-constrained precision agriculture scenarios. In future research, we aim to optimize tensor decomposition algorithms and investigate self-supervised learning strategies to further enhance model generalization and reduce dependency on labeled data.

ACKNOWLEDGMENT

This work is supported by Zhejiang Province 14th Five-Year Plan Teaching Reform Project (Grant No.jg20230464) and Scientific Research Fund of Zhejiang Provincial Education Department (Grant No.Y202456297)